\pdfoutput=1

\documentclass[11pt]{article}

\usepackage[]{EMNLP2023}

\usepackage{times}
\usepackage{latexsym}

\usepackage[T1]{fontenc}

\usepackage[utf8]{inputenc}

\usepackage{microtype}

\usepackage{microtype}

\usepackage{inconsolata}
\usepackage{booktabs}
\usepackage{url}
\usepackage{linguex}
\usepackage{babel}
\usepackage{natbib}
\usepackage{circuitikz}
\usepackage[useregional]{datetime2}
\usepackage{graphicx} 
\usepackage{amsmath}
\usepackage{tcolorbox}
\usepackage{amsfonts}
\usepackage{amssymb}
 \usepackage{mathtools} 
\usepackage{caption}
\usepackage{subcaption}
\usepackage{dsfont}
\usepackage{mathtools}
\usepackage{bbm}
\usepackage{bm}
\usepackage{multirow}
\usepackage{multicol}
\usepackage{natbib}
\usepackage{tikz}
\usepackage{amssymb,amsmath,linguex,amsthm}
\theoremstyle{definition}

\newcommand{\hidden}[1]{}

\let\emptyset\varnothing
\usepackage{graphicx} 

\title{Llamipa: An Incremental Discourse Parser}

\author{Kate Thompson*$^{\dagger}$, Akshay Chaturvedi$^{\dagger}$, Julie Hunter*, Nicholas Asher$^{\dagger\ddagger}$\\ 
       $^{\dagger}$IRIT, $^{\ddagger}$CNRS, *LINAGORA Labs \\ Toulouse, France }

\begin{document}

\maketitle

\begin{abstract}
This paper provides the first discourse parsing experiments with a large language model (LLM) finetuned on corpora annotated in the style of SDRT \citep[Segmented Discourse Representation Theory][]{asher:1993,asher:lascarides:2003}.  The result is a discourse parser, Llamipa (Llama Incremental Parser), that leverages discourse context, leading to substantial performance gains over approaches that use encoder-only models to provide local, context-sensitive representations of discourse units. Furthermore, it can process discourse data incrementally, which is essential for the eventual use of discourse information in downstream tasks. 
\end{abstract}

\section{Introduction}

Understanding a conversation requires more than just grasping the contents of its utterances. One must also understand how different utterances \textit{relate} to one another, for example whether an utterance answers a question or serves to explain or correct something previously said.
Discourse parsing, the task of predicting such  relations, often represents the discourse structure of texts or conversations as graphs in which \textit{elementary discourse units} or EDUs (roughly, clause-level contents)  are represented by nodes, and semantic relations between clauses are represented as labeled edges. 

Discourse parsing remains a challenge in NLP due to three factors: (i) the complexity of discourse graphs, (ii) the frequent lack of surface cues provided by EDUs, which forces parsers to rely on deep, semantic information, (iii) the context dependent nature of the inferences required to predict the graph elements.


Recently, improved discourse parsers have incorporated neural models and encoder-only pretrained language models like BERT \cite{devlin2018bert, bennis:etal:2023} which offer richer representations of linguistic content, and can address factor (ii). But given their limited input size, they are confined to look at one small part of a discourse at a time, and so cannot really address factor (iii).

Generative large language models (LLMs) can substantially improve discourse parsing.  They not only encode large amounts of lexical and conceptual information, addressing problem (ii), but their increased context windows permit an incremental parsing strategy to address problem (iii). Such a strategy makes sense cognitively and linguistically, as prior discourse structure has an effect on inferences about where subsequent discourse units are attached (the link task), and how they are attached, i.e., which types of semantic relations link them (the link+relation task) ~\cite{asher:lascarides:2003}.  Further, an incremental approach naturally applies to downstream applications
such as conversational agents, in which the information about the previously inferred discourse structure must be available and updated in real time. 


We develop an incremental parsing model by fine-tuning Llama3 \cite{dubey2024llama}. {\em Llamipa} (Llama Incremental Parser) uses previously inferred discourse structure  to simultaneously predict the links and relation types between new discourse units. We show that Llamipa substantially outperforms the state-of-the-art on both the link and link+relation tasks.  

We demonstrate Llamipa's performance using three datasets annotated in the style of SDRT \citep[Segmented Discourse Representation Theory][]{asher:1993,asher:lascarides:2003}. Our central dataset is the Minecraft Structured Dialogue Corpus \citep[MSDC;][]{thompson:etal:2024}, a corpus of situated, chat-based, task-oriented dialogues, but we also show the generality of our approach by separately fine-tuning on STAC, a corpus of situated multiparty chats from an online game \cite{asher:etal:lrec}. We further test Llamipa fine-tuned on STAC on an out of domain dataset, Molweni~\cite{molweni}.

Not only does Llamipa outperform previous discourse parsing models, but it handles particularly complex structural features of our situated, multiparty datasets, addressing problem (i). In particular, Llamipa can identify cases in which a single EDU has two parent EDUs, a situation that arises naturally in both multiparty and situated dialogue, where a speaker can respond to multiple channels (conversational or multimodal) at once. Most current SDRT based parsers cannot handle such structures~\cite{shi2019deep,liu-chen-2021,wang:etal:2021,chi:rudnicky:2023}.


In what follows,  Section \ref{sec:discourse-background} provides background on discourse structure. Section \ref{sec:related} presents related work, while Section \ref{sec:corpora} details  the multiparty corpora we use for training and testing. Sections \ref{sec:model} and \ref{sec:results} describe our model and results. Section \ref{sec:ablation} describes our ablation experiments, and we conclude in Section \ref{sec:conclusions}.

\section{Background}\label{sec:discourse-background}
Discourse representations for a text or conversation often take the form of tree or graph structures that are built recursively from discourse units or EDUs, where EDUs are clauses or subclausal units that serve as minimal, linguistic constituents in \textit{discourse relations} \cite{marcu-1999-decision}.  These relations, such as \textit{Explanation}, \textit{Question-Answer Pair} or \textit{Acknowledgement}, are represented as linking EDUs in a discourse graph.  

There are two main theories that have investigated complete discourse structures for texts: Rhetorical Structure Theory \citep[RST;][]{mann:thompson:1987} and SDRT \citep{asher:1993,asher:lascarides:2003}.  RST assigns relations between contiguous EDUs which are connected hierarchically to further EDUs in a treelike structure. 

SDRT provides a dependency style, weakly connected directed acyclic graph (DAG) in which typed relations are represented as labeled arcs in the graph, and the terms of the relations are EDUs or {\em elementary event units} (EEUs). EEUs represent salient aspects of the nonlinguistic context in which a conversation takes place, and may be connected to the discourse DAG.  SDRT also allows subgraphs, or {\em complex discourse units} (CDUs) to act as terms of semantic relations. SDRT is the only framework to date that handles multiparty conversation.

Whatever the underlying paradigm, discourse parsing requires addressing three tasks. First, the segmentation task, which involves decomposing a text and (descriptions of) actions into EDUs and EEUs.  The link task involves identifying pairs of EDUs and/or EEUs that are semantically related via discourse relations---that is, the arcs in the discourse graph. Finally, the link+relation task involves identifying the labels for each attached pair. 

\section{Related work}\label{sec:related}

Traditionally, SDRT style parsers assume gold discourse unit segmentation \cite{muller-etal_2012_papers,afantenos:etal:2015,perret:etal:2016,shi2019deep,liu-chen-2021,chi:rudnicky:2023}, and parsing follows a pipeline architecture in which discourse units are first linked, and then labels are predicted for those links. Our proposed architecture assumes gold discourse unit segmentation, but combines the attachment and labeling tasks.

Several strategies have evolved to harness the complex information required for discourse parsing within the dependency based paradigm proposed by SDRT or DLTag \citep{forbes2003d}.  Research has shown that semantic features of EDUs are crucial for solving the link+relation task \cite{mann:thompson:1987,asher:lascarides:2003}.  
Thus, older approaches to multiparty conversation \cite{afantenos:etal:2015,perret:etal:2016}
input manually designed features of the EDU pairs to simple ML models to predict link and link+relation label successively.  These \textit{local models} use information local to the pairs of EDUs that are to be attached, but do not exploit information about the surrounding discourse structure. In order to leverage some more global information, \citet{afantenos:etal:2015,perret:etal:2016} apply special decoding mechanisms with post hoc constraints on the output EDU pairs, following~\citet{muller-etal_2012_papers}.  Although local models exploit little to no information about other components of the discourse structure, such a model can still be competitive~\cite{bennis:etal:2023}.  We do a detailed comparison between an encoder based local parsing model and Llamipa.

\citet{shi2019deep} were the first to obtain competitive discourse parsing results using a deep learning architecture on STAC. They attempted to capture incremental and contextual effects in their model by training a supplemental \textit{Structured Encoder} to incrementally update attachment paths (sequences of parent-child EDUs). 
 However, \citet{wang:etal:2021} show that the model obtains similar scores with or without this encoder. 

\citet{liu-chen-2021} develop an approach that uses a pre-trained RoBERTa model~\cite{roberta} to provide embeddings for EDUs.  In addition, they use a bi-GRU cell~\cite{cho-etal-2014-properties} to capture contextual information; but this requires them to limit the size of the input.  Their model uses two linear layers for link and relation prediction. 


\citet{wang:etal:2021} provide the  Structure Self-Aware Graph Neural Network (SSA-GNN), which is a complex GNN-based architecture and model that uses both EDU and relation embeddings. 
The model uses a Hierachichal GRU gate for contextual EDU representations. They then apply the SSA-GNN to capture implicit structural information between EDUs, using a Structure-Aware Scaled Dot-Product Attention \cite{zhu:etal:2019,wang:etal:2020} to update edge and EDU representations. They further train a teacher network to supplement the standard classification loss with an auxiliary loss. \citet{chi:rudnicky:2023} developed a more complex model based on \citet{shi2019deep}.

None of the models by \citet{shi2019deep,liu-chen-2021,guz:etal:2020, wang:etal:2021,chi:rudnicky:2023} predict multiple parents for an EDU (or EEU).  
The only previously developed parser that predicts multiple parents during attachment is \citet{bennis:etal:2023}'s model BERTLine.   BERTLine uses BERT embeddings for EDUs and EEUs, along with a linear layer for features like speaker change and distance information, to build a local model and predicts link and link+relation in pipeline fashion. However the relation labeling uses a multitask architecture to capture informational dependencies between link and relation labeling decisions.   ~\citet{bennis:etal:2023} show that their model is competitive with approaches that try to build in more contextual knowledge \cite{liu-chen-2021,wang:etal:2021}.  \citet{thompson:etal:2024} uses BERTLine on the MSDC, which makes it a natural baseline for our work on situated, multiparty conversation.  

Recent work investigates using LLMs to predict discourse structure by finetuning Llama2 7b, 13b, and 70b for RST-style parsing \cite{maekawa:etal:2024}. The reported results show  improvements over the state-of-the-art for RST corpora analogous to those we find using Llamipa on SDRT corpora.  \citet{chan:etal:2024,yung:etal:2024,fan-etal-2024-uncovering-potential} have investigated discourse parsing through prompting with large GPT-style models to produce PDTB \citep{prasad:etal:pdtb}  or other discourse annotations.  These models fare worse than dedicated discourse parsers: \citet{chan:etal:2024} test with 3-shot learning on a version of the STAC corpus (see Section \ref{sec:corpora} for more corpora details) and report a near 50 point drop in F1 for link and link+relation prediction compared to \citet{shi2019deep}.


\section{Data sets}\label{sec:corpora}

\textbf{MSDC.} The MSDC provides full discourse structures in the style of SDRT for dialogues in the Minecraft Dialogue Corpus \citep[MDC;][]{narayan-chen-etal-2019-collaborative}.  
\begin{figure}[!ht]
\begin{center}
\includegraphics[scale=0.42]{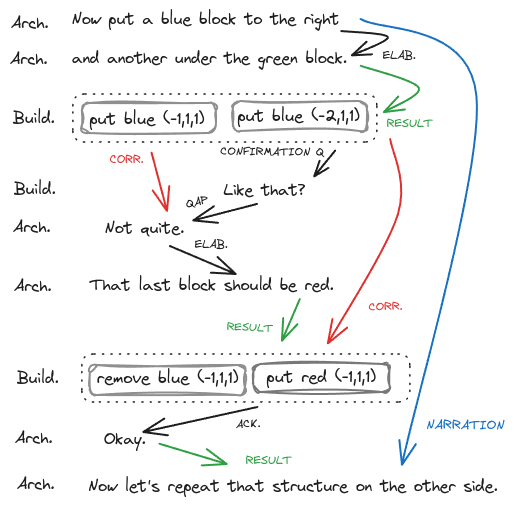} 
\caption{MSDC dialogue example showing CDUs composed of EEUs for builder action sequences and multiparent discourse units (MPDUs).   The Narration on the right links ``high level'' instructions, which are particularly hard to predict, as explained in Section \ref{sec:ablation}.}
\label{fig:correction}
\end{center}
\end{figure}

The MSDC represents conversational moves as well as builder move sequences, individuated into block placements and removals. Figure \ref{fig:correction}
shows an excerpt from an MSDC dialogue containing a few examples of discourse units with multiple parents, or MPDUs. In one such example, after placing two blue blocks, the builder asks, ``like that?''. The architect's reply, ``not quite,'' not only serves to answer the builder's question, but also corrects the builder's last moves. The EDU for ``like that?'' is an MPDU that is semantically linked to both linguistic and nonlinguistic units. Such MPDUs are an important feature of the MSDC, as they facilitate the interplay of multimodal contributions in a single discourse structure.  


Another noteworthy feature of the MSDC is how nonlinguistic actions are annotated. Builder moves are usually prompted by architect instructions, but often transpire in longer sequences of block placements and removals, which sometimes overlap with other linguistic chat moves. These action sequences are grouped together in single complex discourse units (CDUs) in the MSDC, indicated by the dashed boxes in Figure \ref{fig:correction}. \citet{bennis:etal:2023}, \citet{thompson:etal:2024} merged these highly repetitive sequences of nonlinguistic moves into a single EEU to prevent the repetitive connections between them from drowning out the more complex and interesting linguistic interactions in the corpus.  We used the MSDC with this modification. 



\begin{table} 
\centering
\small{
\begin{tabular}{llll}
\hline
\textbf{DU-type} &\textbf{STAC-L} & \textbf{STAC-Sit} & \textbf{MSDC}  \\
& Train/Test & Train/Test & Train/Test\\
\hline\\
EDU & 11398/1154 & 11707/1163 & 17135/5402 \\
EEU  &0/0 & 11566/1156 & 4687/1473\\
MPDU  & 838//86& 999/96 & 4789/1476\\
MPDU=3 & 94/9 & 95/6 & 85/35\\
MPDU>3 & 11/0 & 7/0 & 1/3\\
\hline
\end{tabular}}
\caption{\label{tab:stats} EDU, EEU, and MPDU counts for SDRT datasets.  MPDU-3 gives the number of MPDUs with 3 parents, MPDU>3 those with more than 3 parents.}
\end{table}


\textbf{STAC}. STAC\footnote{\url{https://https://www.irit.fr/STAC/corpus.html}} is a corpus of multiparty chats from the game \textit{Settlers of Catan}, in which players trade or otherwise acquire resources in order to build roads and settlements and thereby score victory points.   STAC has two versions: \textit{STAC-L}, in which only relations between chat moves (EDUs) are annotated in the absence of the context provided by nonlinguistic events, and a situated version, {\em STAC-Sit}, in which both chat moves and (descriptions of) game moves (EEUs) are annotated in the presence of an integrated linguistic and nonlinguistic context. 

Like MSDC, STAC contains MPDUs due to multimodal interactions, but also features MPDUs which arise from multiparty interaction, in which one player can simultaneously respond to multiple players at once. Relevant statistics for STAC and the MSDC are found in Table \ref{tab:stats}.


STAC-Sit, like the MSDC, also contains sequences of moves containing only EEUs. \citet{bennis:etal:2023} compressed these sequences into individual EEUs and we follow this choice. Unlike the MSDC, however, STAC contains numerous CDUs for chat moves and for chat/game interactions (as opposed to CDUs of action sequences only). 
Following \citet{bennis:etal:2023} and most of the literature on SDRT parsing, we do not treat CDUs directly, but use the ``flattening'' procedure outlined in \citet{muller-etal_2012_papers} that converts a structure with CDUs into one with only EDUs and EEUs. A final simplification of the original STAC corpus involved removing EEUs that neither interact with EDUs nor figure in EEU sequences that interact with EDUs. This choice, like the one to compress EEU sequences, is made to avoid drowning out the more complex and interesting data involving linguistic EDUs and their interactions with EEUs.  
  

\textbf{Molweni}. The Molweni corpus contains SDRT graphs for multiparty interactions in the Ubuntu Chat Corpus \cite{lowe-etal-2015-ubuntu}, a collection of exchanges from an online forum for discussing issues related to Ubuntu. Unlike the MSDC and STAC, Molweni does not contain CDUs, EEUs, or MPDUs.  We use the Molweni-clean test set, a version of Molweni re-annotated by \citet{li:etal:2024}.

\section{Llamipa: An Incremental Parser} \label{sec:model}
Llamipa leverages the enhanced contextual reasoning capacities of generative LLMs by introducing an {\em incremental discourse parsing strategy}. As it generates a discourse graph for a conversation, the LLM bases its link and relation predictions for an elementary unit $\epsilon$ based on the graph it has already constructed for units preceding $\epsilon$. 

Formally, let $C$ be a sequence $\epsilon_0,..., \epsilon_n$ where $\epsilon_i$ denotes an EDU or EEU. Our LLM takes as input a set $U$, containing EDUs and EEUs, and a discourse graph $G$. At step $0$, $U$ is $\{\epsilon_0\}$, the head EDU, and $G$, the null discourse graph. After that, $U$ is a set $\{\epsilon_l,..., \epsilon_{m}\}$, where $0 \leq l < m \leq n$ and $m-l \leq k$ where $k$ is the window size, and $G$ contains links connecting $\{\epsilon_l,..., \epsilon_{m}\}$. At each step, the model extends the previous discourse graph by integrating link and relation predictions for all of the EDUs or EEUs in the current dialogue turn.


\begin{figure}[t]
    \centering
    \includegraphics[scale=0.135]{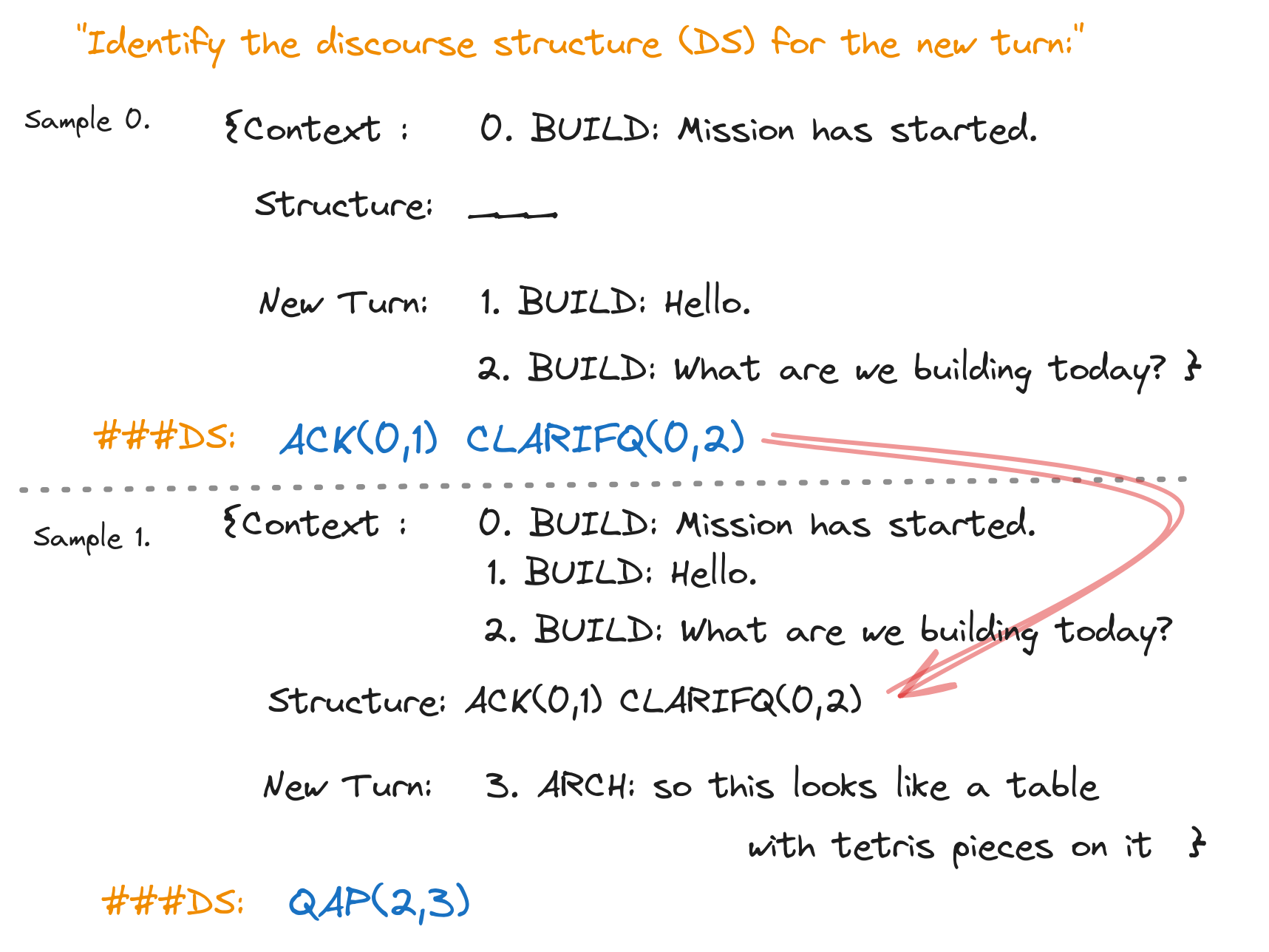}
    \caption{Depiction of dialogue increments seen during Llamipa3+p generation. During generation, the predicted relations are added to the context structure for the following increment.}
    \label{fig:data}
\end{figure}
In the data, we represent relation instances as elementary unit index pairs connected by a relation type, e.g. \textit{RES(0,1)}, for the instance of a \textit{Result} relation holding between units $0$ and $1$.  During training, each sample contains the previous EDUs and EEUs within the window size, the gold structure, which is a list of relation instances between these elementary units, and finally, the units of the current turn. The model is trained to predict the gold relation instances that connect the units of the current turn with previous ones.


As LLMs can handle longer input sequences, we use a window size $k = 15$ which allows the model to exploit a large amount of contextual information. 
Since discourse links follow an inverse logarithmic curve \cite{afantenos:etal:2015}, there is little statistical support for links of distance longer than $15$ in the SDRT corpora.

We finetune Llama 2 \cite{touvron2023Llama}, and Llama 3 \cite{dubey2024llama}
to produce Llamipa2 and Llamipa3 using QLoRA \cite{dettmers2024qlora}. Both the models are trained for 3 epochs. The model parameter details are in Table \ref{tab:model-details} in Appendix \ref{sec:appendix-a}. We release Llamipa3 publicly through the Hugging Face platform\footnote{\url{https://huggingface.co/linagora/Llamipa}}.

We test our models in two settings. In the first setting, the models predict the relation instances for samples containing the gold discourse structure. In the second, the model uses its predicted relation instances instead of the gold structure in the input sample. That is, at each step, the structure fed to the model is just the accumulated relation instances predicted within the window up until that step (see Figure~\ref{fig:data}). This incremental setting provides a clear idea of the model's ability to build a complete dialogue structure from scratch.  

\section{Results}\label{sec:results}
We run finetuning experiments using the MSDC corpus, and assess Llamipa's performance on link and link+relation prediction under the two test settings described in Section~\ref{sec:model}. We consider only the well-formed outputs during evaluation, but note that Llamipa outputs only well-formed relation instances for all experiments. 

\begin{table}
\centering
\small{
\begin{tabular}{@{}lllll@{}}
              & \multicolumn{2}{l}{MSDC} & \multicolumn{2}{l}{STAC Situated} \\ \midrule
              & Link     & Link+Rel     & Link        & Link+Rel       \\ \toprule
Llamipa2+g   & 0.8561       & 0.7664        & -           & -           \\
Llamipa2+p   & 0.8579       & 0.7570        & -           & -          \\
Llamipa3+g   & 0.9004       & 0.8154        & -          & -          \\ 
\textbf{Llamipa3+p}   & \textbf{0.8830}       & \textbf{0.7951}       & \textbf{0.8612}          & \textbf{0.7796}          \\\midrule
BERTLine      & 0.7870       & 0.6901        & 0.7667          & 0.6788          \\ \midrule
\end{tabular}
}
\caption{\label{tab:lamipa}
Performance of Llamipa2 and Llamipa3 on the MSDC test set for link prediction, evaluated using micro-averaged F1, and link+relation prediction, evaluated using a weighted average F1. We show results for Llamipa using gold structure (g) and  predicted structure (p)  as input. We also show link and link+relation results for Llamipa3+p finetuned on STAC-Sit and evaluated on STAC-Sit test with predicted inputs. All the results are for relation distance $\leq$ 10 EDUs, the cutoff used by BERTLine.}
\end{table}

\begin{table*}
\centering
\begin{tabular}{@{}lllll@{}}
& \multicolumn{2}{l}{STAC-L} & \multicolumn{2}{l}{STAC- L/Molweni-clean} \\ \midrule
                        & Link           & Link+Rel          & Link               & Link+Rel              \\ \toprule
\citet{perret:etal:2016}            & 0.69             & 0.531             & -                    & -                    \\
\citet{shi2019deep}               & 0.719             & 0.536             & -                    & -                    \\
\citet{liu-chen-2021}              & 0.731            & 0.571             &  -                  & -                   \\
\citet{wang:etal:2021}             & 0.734             & 0.573             & -                    & -                    \\
\citet{fan2022distance}& 0.736             & 0.574             & -                    & -                    \\
\citet{chi:rudnicky:2023}           & 0.744             & 0.596             & 0.645                 & 0.380                 \\ 
\citet{li:etal:2024}                & 0.593             & 0.386             & 0.756                & 0.312                 \\ 
\citet{bennis:etal:2023} (BERTLine) & 0.730             & 0.562             & -                    & -                    \\\midrule
\citet{fan-etal-2024-uncovering-potential} (ChatGPT)           & 0.599             & 0.252             & -                    & -                    \\
\citet{chan:etal:2024} (ChatGPT)             & 0.213             & 0.074             & -                    & -                    \\ \midrule
Llamipa-3+p                   & 0.775           & 0.607             & 0.712                 & 0.405                
\end{tabular}
\caption{\label{tab:comparison}
A comparison of the performance of various models on  STAC-L, the linguistic version of the STAC dataset, as well as out-of-domain performance on a re-annotated version of the Molweni test set provided by \citet{li:etal:2024}. 
Our STAC-L is a flattened version of STAC-L in the STAC repository.  See Appendix \ref{sec:appendix-b} for discussion of STAC-L possible variants.}
\end{table*}

Table \ref{tab:lamipa} shows that Llamipa3 significantly outperforms BERTLine and outperforms Llamipa2 both using gold structure (Llamipa3+g vs. Llamipa2+g) and predicted structure (Llamipa3+p vs. Llamipa2+p). 

To test the generalizability of our approach, we also finetune Llamipa on STAC-Sit and compare the Llamipa3+p outputs to BERTLine. Llamipa3+p shows a gain of $10$ F1 points on link prediction and $10.5$ F1 points on link+relation prediction. As mentioned in Section~\ref{sec:model}, the predicted context setting best demonstrates the desired functionality of an incremental parser. Since Llamipa3+p give the best performance in this setting, we use it for all evaluations and henceforward refer to it using the shorthand \textit{Llamipa}.

\textbf{Performance on STAC-L}. 
In order to compare Llamipa to a broader range of SDRT discourse parsers, we tested it on STAC-L, as prior work has only used this linguistic version of STAC. Table \ref{tab:comparison}  shows that Llamipa finetuned on STAC-L and evaluated on the STAC-L test set scores substantially higher on both link and link+relation prediction than the state-of-the-art parser \cite{chi:rudnicky:2023}. 
We note that Llamipa has a lower performance on STAC-L than on either of the situated corpora. This reflects the findings of \citet{asher:etal:lrec,asher:etal:2021}, who reported that annotating the STAC linguistic data set was more difficult; annotators felt that the nonlinguistic information needed to determine the discourse structure was missing (see Section~\ref{sec:corpora}).

\begin{table*}
\centering
\small
\begin{tabular}{lllllllllllllll}
\hline
\textbf{Distance} &
2& 3 & 4 & 5 & 6 & 7 & 8 & 9 & 10 &11 &12 & 13 & 14 & 15\\
\hline
\textbf{Llamipa-3}  &0.90 & 0.88 & 0.82 & 0.76 & 0.82 & 0.80 & 0.79 & 0.71 & 0.90 & 0.76 & 0.50 & 0.40& 1.0 & n/a\\
 \textbf{BERTLine}
 &0.70 & 0.61 & 0.43 & 0.32 & 0.38 & 0.02 & 0.21& 0.28 &  0.08 &n/a & n/a & n/a & n/a & n/a \\

\hline
\end{tabular}
\caption{\label{tab:narrations}
F1 score for Llamipa and BERTLine on Narration instances in the MSDC test with distance between 2 and 15 EDUs. There were 0 instances of Narration at distance 15 in gold.  BERTLine's input was limited to pairs whose distance was $\leq$ 10.}
\end{table*}
 
 We also test Llamipa on transfer learning. Following \citet{li:etal:2024}, we finetune on STAC-L and test on their Molweni-clean test set. While Llamipa outperforms all models on STAC-L, it fares worse on link prediction than \citet{li:etal:2024} on Molweni, though it does better on link+relation prediction; a counterintuitive result given that link+relation prediction is the harder task. Taking a closer look at the link prediction results in \citet{li-etal-2023-discourse}, we see their model performs well on links of distance $1$, but drops after that ($\sim 25\%$ accuracy for distance $2$, and $0$ for distances greater than $5$). Given that Molweni is a relatively simple corpus where most of the links are adjacent, their model excels on the link task. But for relation type prediction, \citet{li:etal:2024} use the traditional method, which assigns relation types after establishing the attachment structure, and using only local information. Llamipa surpasses their model on relation type prediction, which shows that an approach predicting link and relation simultaneously performs better than a local pipeline approach, even for shallower discourse structures. 

\textbf{Performance on MSDC}. 
Llamipa shows impressive gains for predictions of certain relation types. Looking at F1 breakdown by relation type in Table \ref{tab:rel-type}, we see the largest relative increase comes from improved predictions for \emph{Narration} and \emph{Correction}, which \citet{thompson:etal:2024} cite as the most difficult to predict given that they are heavily context dependent.
\textit{Narrations} are frequently quite long (with a significant proportion  above distance $10$), and difficult to capture by local parsers like BERTLine~\cite{thompson:etal:2024}. 
Table \ref{tab:narrations} shows Narration F1 at distances between $2$ and $15$ EDUs. Like other local parsers, BERTLine requires a distance cutoff that balances ignoring longer relations with mitigating the class imbalance caused by numerous unlinked candidates; for SDRT corpora, this is distance $10$. As indicated in Section \ref{sec:model}, Llamipa continues to make good predictions up to distance $15$.  

Further, Llamipa is able to predict many instances of the \textit{Correction} relation, which also has a number of long distance instances in the MSDC. Table \ref{tab:rel-type} shows that its performance is $73$ F1 points, which is 42 points higher than and more than double that of BERTLine.  

We hypothesize that Llamipa has a high F1 on \textit{Narration} and \textit{Correction} because it has access to rich semantic information about the discourse context: it can use the previous discourse moves as well as an explicit representation of their discourse structure.  This hypothesis coheres with our understanding of how \textit{Narration} functions in the MSDC. \textit{Narrations} connect high-level instructions between which participants clarify, question and elaborate on those instructions, as well as correct missteps (see Figure \ref{fig:correction}). While there is some statistical similarity in their surface forms, the important cues for isolating a new instruction, and hence a \textit{Narration} attachment, involve certain patterns of \textit{Result} and \textit{Acknowledgement} links in the prior context, as well as the presence of (nonlinguistic) block placements. This pattern is regular enough that one can find most \textit{Narrations} programmatically as with \citet{thompson:etal:2024}'s ``Second Pass''. This takes predictions from a first pass of the local model for \textit{Results} and \textit{Acknowledgements}, then applies an algorithmic approach to predict long distance Narration links. 
We tried this second pass for Llamipa, however there was no improvement on F1 for Narration scores, indicating that Llamipa has already internalized this pattern during training. 

Llamipa offers improvements over BERTLine for many other relation types which are typically not considered long distance.  Table \ref{tab:rel-type} shows Llamipa has higher scores for \textit{Result}, \textit{Acknowledgment}, \textit{Elaboration}, \textit{Question-answer Pair}, \textit{Clarification Question}, and \textit{Comment}.  We hypothesize that predictions on these also profit from complex contextual information, something that we discuss in more detail in Section~\ref{sec:ablation}. Conversely, for relations where Llamipa gives slightly poorer results---\textit{Contrast}, \textit{Alternation}, and \textit{Question-Elaboration}---we surmise that these types can be inferred reliably just from the local information given by the related EDUs or EEUs.  

Overall, Llamipa has several advantages.  It simplifies previous attach-then-type pipeline methods by predicting link and relation labels simultaneously. Previous research has given theoretical arguments that co-predicting attachment and relation labels should give better results \cite{asher:lascarides:2003}, as well as empirical evidence for this claim~\cite{bennis:etal:2023}.  And as we have demonstrated, Llamipa  outperforms other parsers on all relevant corpora. 

\begin{table*}
\centering
\small{
\begin{tabular}{@{}lllllll@{}}
    & BERTLine & Llama3-local & Llamipa3+p & Llamipa3+g & Llamipa+rand & Llamipa+$\emptyset$ \\ \toprule
Result        & 0.85     & 0.86    & 0.90      & 0.91     & 0.83         & 0.72  \\
Acknowledgement & 0.81     & 0.85   & 0.86      & 0.86     & 0.81         & 0.72    \\
\textbf{Narration}      & \textbf{0.50} [0.73]     & \textbf{0.54}    & \textbf{0.82 } & \textbf{0.91} & \textbf{0.28}  & \textbf{0.18}    \\
Elaboration     & 0.75     & 0.77   & 0.77      & 0.77     & 0.75         & 0.68   \\
Correction     & 0.31     & 0.64     & 0.73      & 0.80     & 0.49         & 0.52    \\
Continuation    & 0.44     & 0.47   & 0.49      & 0.50     & 0.44         & 0.33   \\
Question-answer Pair  & 0.76  & 0.80  & 0.82  & 0.83 & 0.42  & 0.14     \\
Comment     & 0.50     & 0.57         & 0.60   & 0.61     & 0.57         & 0.54    \\
Confirmation-Question    & 0.86  & 0.89  & 0.93  & 0.93     & 0.91     & 0.89  \\
Clarification-Question   & 0.61   & 0.66 & 0.73   & 0.73     & 0.70    & 0.41   \\
Contrast    & 0.80     & 0.79     & 0.75  & 0.74   & 0.72     & 0.68     \\
Question-Elaboration   & 0.39     & 0.36  & 0.30      & 0.36     & 0.33     & 0.33    \\
Alternation    & 0.88     & 0.88          & 0.83      & 0.88     & 0.90     & 0.96     \\
Explanation   & 0.00     & 0.00          & 0.00      & 0.00     & 0.00         & 0.00    \\
Conditional   & 0.58     & 0.00          & 0.00      & 0.00     & 0.00         & 0.00     \\
Sequence   & 0.00     & 0.00          & 0.00      & 0.00     & 0.00         & 0.05    \\ \midrule
Link+Rel F1 & 0.69 [0.71]     & 0.73          & 0.80       & 0.81     & 0.65         & 0.56  \\
 Link F1     & 0.78 [0.80]   & 0.82          & 0.88     & 0.90     & 0.77         & 0.72        
\end{tabular}
}
\caption{\label{tab:rel-type}
Performance breakdown for BERTLine, Llamipa2 and Llamipa3, Llama3-local by relation type on MSDC. Relation types are listed in descending order from most to least number of instances in the corpus.  The [F1] scores for Narration after the Second Pass on BERTLine  as reported in \citet{thompson:etal:2024} shown in brackets. The three columns on the right show the effects of varying the context structure during Llamipa's generation step. }
\end{table*}

\section{The role of context}\label{sec:ablation}

So far, the hypothesis has been that Llamipa's state-of-the-art performance, especially for longer distance relations, is due to its ability to process richer contextual information. Whether this is a function of the superior embeddings conferred by the underlying LLM, or a function of a larger context window that allows a longer dialogue history and explicit discourse structure as prediction input, is something we explore further in this section.


\textbf{First ablation: embeddings vs. context size}. 
 This ablation tests whether the LLM embeddings for EDUs and EEUs account for Llamipa's performance. We finetune Llama3-8b over pairs of elementary units as input to predict links and relation types, resulting in a local model roughly comparable to BERTLine. Table \ref{tab:rel-type}
shows that, compared to BERTLine, Llama3-local still performed better without using the prior discourse context for link and link+relation prediction. However, when compared to Llamipa, Llama3-local still has considerably lower predictions overall. Llamipa's relation predictions for many relations are also better, notably for \textit{Narration}. This supports our conclusion that although Llama's superior embeddings play a role, many relations, and especially relations like \textit{Narration}, require the information available in Llamipa's larger context window.

\textbf{Second ablation: perturbing input structure}. 
Even if Llamipa requires the input available in a larger context window, it is still an open question whether it uses the discourse structure provided in this window.  

We already have some indication that Llamipa \textit{does} use the structure when we look at  Llamipa's performance given a gold discourse structure for a context versus its predicted structure, as described in Section~\ref{sec:model}. The results of this comparison are shown in Table \ref{tab:rel-type}: there is a small but consistent drop between gold and predicted structure contexts in F1 performance for all relation types. 

To investigate this further, we perform a second ablation where we perturb the input structure for each test sample by randomizing both the relation type and EDU index for each relation instance in the structure. When we run Llamipa on this randomized input (denoted as Llamipa+rand in Table \ref{tab:rel-type}) we see a drop in performance from Llamipa3+p for link and link+relation for most relation types, with a substantial drop for \textit{Narration}, \textit{Correction}, and \textit{Question-answer Pair}. 

\begin{figure*}[t]
    \centering
    \includegraphics[width=\textwidth,height=4cm]{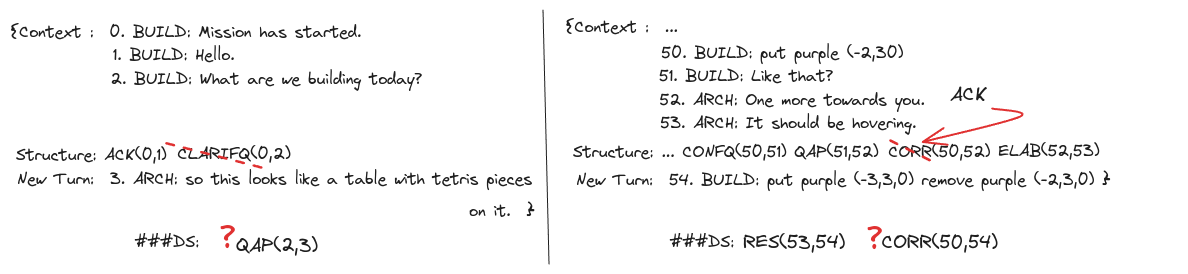}
    \caption{Depiction of targeted perturbation ablations described in Section~\ref{sec:ablation}. Left: We remove the \textit{Clarification-question} from the structure of samples that should predict \textit{Question-answer pair relations}, and see a 50 point drop in F1 for \textit{Question-answer pair} prediction. Right: We change the first \textit{Correction} in the structure to \textit{Acknowledgement} in samples that should predict a second \textit{Correction} in response to the first.}
    \label{fig:ablation}
\end{figure*}

\textbf{Third ablation: removing input structure}. In this ablation, we removed all structure from Llamipa's input context, leaving only a sequence of elementary units. The performance of Llamipa on this data set (in the column Llamipa+$\emptyset$ of Table \ref{tab:rel-type}) shows an even larger drop in both link and link+relation predictions compared to Llamipa+rand.  Similar to Llamipa+rand, we see a substantial drop for the relation types \textit{Narration}, \textit{Correction}, and \textit{Question-Answer Pair}. Additionally, the performance dropped drastically for \textit{Clarification Questions}. 

\textbf{Fourth ablation: targeted perturbations.}
The results of the previous ablations  support the hypothesis that the structure provided in the context window \textit{does} have an effect on relation instance predictions. This final ablation investigates whether Llamipa predicts context dependent relations as well as it does because it leverages (or ``reasons'' about) the relation instances in the context structure in a manner consistent with our semantic intuitions. Here, we focus on instances of \textit{Question-answer pair} (\textit{QAP}) and \textit{Correction}. 

\textit{QAP} attachments provide a straightforward example of context dependence. 
In the MSDC, it is common that if a prediction \textit{QAP(y,z)} is made for a dialogue turn \textit{z} then \textit{CLARIFQ(x,y)}, \textit{CONFQ(x,y)}, or \textit{QELAB(y,z)} is in its context. That is, the questions that are answered in the QAP relations tend to be questions asked to clarify, confirm or elaborate upon the preceding  context. To test whether Llamipa's predictions conform to this pattern, we isolate the test samples where Llamipa correctly predicts \textit{QAP} (Table~\ref{tab:rel-type}), and remove all question relation types from the structure given in the context of these samples. We re-run the generation step on this amended sample set, and see a  $50$ point drop in F1, from $0.87$ to $0.37$, for \textit{QAP} predictions, and no change for all other relation types (see Figure~\ref{fig:ablation}).

\begin{figure}[!ht]
\centering
\small
\resizebox{0.4\textwidth}{!}{%
\begin{circuitikz}
\tikzstyle{every node}=[font=\scriptsize]
\draw[rounded corners]  (5,16.8) rectangle  node {instruction I} (6.4,16.5);
\draw[rounded corners]  (4.9,16.2) rectangle  node {action X} (6,15.8);
\draw[rounded corners]  (3.75,15.2) rectangle  node {instruction Y} (5.15,14.9);
\draw[rounded corners]  (5.1,14.4) rectangle  node {action Z} (6.2,14);
\draw [->, >=Stealth] (5.5,16.5) -- (5.5,16.2);
\draw [->, >=Stealth] (5.3,15.8) -- (4.6,15.2);
\draw [->, >=Stealth] (5.6,15.8) -- (5.55,14.4);
\draw [->, >=Stealth] (4.6,14.9) -- (5.1,14.2);
\node [font=\tiny] at (4.2,15.6) {CORR(X, Y)};
\node [font=\tiny] at (6.3,15) {CORR(X, Z)};
\node [font=\tiny] at (4.3,14.5) {RES(Y, Z)};
\end{circuitikz}
}%
\caption{Correction triangles are composed of a \textit{Correction} to an action sequence, followed by a new action sequence that is the \textit{Result} of the corrective move and a \textit{Correction} of the first action sequence.}
\label{fig:triangle}
\end{figure}

\textit{Correction} relations are among the most difficult to identify in the MSDC, as discussed in Section~\ref{sec:results}. Although the language used for \textit{Corrections} usually features some form of negation---\textit{No, Not like that, Oops!}----deciding where it attaches requires making inferences about how the content of the corrective move applies to a more global information state. Although this is beyond the capability of a discourse parser like Llamipa, there are still structural regularities in the MSDC that can help with retrieving \textit{Corrections}. 

For instance, most \textit{Corrections} come about when the Builder makes the wrong moves in response to an instruction, and the Architect then gives a further instruction, which  results in moves that address this by fixing the last wrong moves. In the discourse structure, these interactions create recurring patterns, ``correction triangles'' (Figure~\ref{fig:triangle}). These triangles provide a way to check Llamipa's sensitivity to \textit{CORR(x,y)} type relations in the context structure in cases where it must predict \textit{RES(y,z) CORR(x,z)}, the last two relations in a triangle. As with the \textit{Question-answer Pair} test, we single out each test sample in which Llamipa correctly predicts the second \textit{CORR}; but instead of removing \textit{CORR}(x,y), we change it to \textit{ACK}(x,y) 
 ({\textit{Acknowledgement}(x,y)). This leaves the structure intact, but gives it a contrary sense. We re-run the generation step and see a $20$ point drop in F1, from $0.71$ to $0.52$, for \textit{Correction} predictions, and no change for all other relation types.

\section{Conclusions} \label{sec:conclusions}

Llamipa is an incremental SDRT style discourse parser that uses its prior discourse structure predictions to predict links and relation labels for the discourse units in new conversation turns. As such, it can be naturally assimilated into downstream applications like conversational agents. 

After finetuning both Llama2-13b and Llama3-8b models to produce Llamipa2 and Llamipa3, we show that Llamipa3 has substantially higher F1 scores than other parsers that use SDRT data sets on the link and link+relation tasks. We confirm that its performance is due to access to larger swaths of context, which are not available to parsers that use a local model. We further find evidence to support the hypothesis that Llamipa uses the explicit representations of structure in its context to predict new relation instances. 

Notably, Llamipa excels on data sets that feature conversation containing both linguistic and nonlinguistic moves (actions, in the MSDC) that help interpret them. This indicates that a parser like Llamipa can furnish important discourse information to builder action models; preliminary demonstrations of this can be found in \citet{chaturvedi2024nebula}. 


\section*{Limitations}
We did not exhaustively test all models on the situated data sets.  Given the number of MPDUs especially in the MSDC, however, we felt that their performance would not compare to Llamipa's. We plan to do more extensive testing in the future to compare models. 

Llamipa, like previous parsers, presumes that the EDUs/EEUs are already segmented. However, existing work on automatic segmentation suggests that this task is relatively tractable on diverse data sets \cite{muller:etal:tony} including even corpora of spoken conversation \citep{gravellier2021weakly,prevot2023comparing}  


Like all supervised models, Llamipa is limited to making predictions on data sets that are similar to its training data. Given the heavy annotation effort required to build data sets annotated for discourse structure, this arguably puts a bottleneck on the generalizability of our approach to data sets in languages other than English and for conversation styles, such as multiparty spoken conversation, that are significantly different from the chat-based dialogues in the MSDC and STAC. Future work will explore efficient ways of harnessing extant data sources through, for instance, automatic translation, data augmentation and or/heuristics, to facilitate the transfer to other languages and domains.


\section*{Ethics Statement}
Our work here has been to improve the capacities of AI models to understand the semantic structure of discourse and conversation.   We see no direct ethical concerns that arise from this work, though we hope that a better grasp of the semantic structure of conversation may enable the output of models to be more pragmatically grounded.

\section*{Acknowledgements}
For financial support, we thank the National Interdisciplinary Artificial Intelligence Institute ANITI (Artificial and Natural Intelligence Toulouse Institute), funded by the French ‘Investing for the Future–PIA3’ program under the Grant agreement ANR-19-PI3A-000. This project has also been funded by the France 2030 program and is funded by the European Union - Next Generation EU as part of the France Relance. We also thank the projects COCOBOTS (ANR-21-FAI2-0005) and DISCUTER (ANR-21-ASIA-0005), and the COCOPIL ``Graine'' project of the Région Occitanie of France. This work was granted access to the HPC resources of CALMIP supercomputing center under the allocation 2016-P23060. 


\clearpage
\appendix

\section{Appendix: Model Parameters}
\label{sec:appendix-a}





\begin{table}[h]
\begin{tabular}{@{}ll@{}}
\toprule
\multicolumn{2}{c}{GPUs}             \\ \midrule
\multicolumn{2}{c}{4 NVIDIA Volta V100} \\ \midrule\midrule
\multicolumn{2}{c}{Hyperparameters} \\ \midrule
Training epochs             & 3     \\
batch size                  & 4     \\
optimizer                   & Adam     \\
learning rate               & 2e-4   \\
\multirow{2}{*}{learning rate scheduler}     & linear warm-up and     \\
                                            & cosine annealing \\
warm-up ratio               & 0.03    \\
gradient clipping           &  0.3    \\
lora r                      & 64     \\
lora (alpha)                & 16     \\
lora dropout ratio          & 0.1     \\
\multirow{2}{*}{lora target modules}         &  Only Attention Blocks\\ 
                                        & (q\_proj, v\_proj)    \\
quantization for Llama3     & 4-bit NormalFloat \\ \bottomrule
\end{tabular}
\caption{\label{tab:model-details}Details on computing resources and hyperparameters for finetuning Llamipa.}
\end{table}

Table~\ref{tab:model-details} gives the hyperparameters we use for finetuning Llamipa-2 and Llamipa-3 along with the computing resources. We adapt the finetuning code from the following repository\footnote{\url{https://github.com/mlabonne/llm-course/blob/main/Fine_tune_Llama_2_in_Google_Colab.ipynb}}. 

\section{Appendix: STAC Linguistic data}
\label{sec:appendix-b}

We derive STAC-L from the STAC data available on the website (\url{https://https://www.irit.fr/STAC/corpus.html}) by adopting standard procedures for flattening CDUs.  However, we find that our data set did not quite match that of \citet{bennis:etal:2023} in terms of numbers of EDUs.  ~\citet{shi2019deep} also report a different number of EDUs from what we have \textit{and} from what \citet{bennis:etal:2023} report. When we finetune and test Llamipa on the STAC-L data available in the BERTLine repository \footnote{\url{https://github.com/zineb198/LineBert}}, we had lower scores than what we report in Table \ref{tab:comparison}. Llamipa-3+p had an F1 for link of 75.4 and a link+relation F1 of 57.7 on this version of STAC linguistic. On the transfer learning from STAC linguistic to the Molweni-clean test set, Llamipa-3+p had an F1 of 68.5 for link and 38.7 for link+relation. We make our version of STAC-L available on the STAC corpus website cited above.
\end{document}